\documentclass[conference]{IEEEtran}
\IEEEoverridecommandlockouts
\usepackage{fancyhdr}

\fancypagestyle{firstpage}{%
  \fancyhead{}
  
  \fancyfoot[C]{Accepted for Proceedings of IJCNN 2025, Rome, Italy. Copyright IEEE}
}
\usepackage{cite}
\usepackage{float}
\usepackage{amsmath,amssymb,amsfonts}
\usepackage[ruled,vlined]{algorithm2e}
\usepackage{graphicx}
\usepackage{gensymb}
\usepackage{textcomp}
\usepackage{xcolor}
\usepackage{steinmetz}
\usepackage{balance}
\def\BibTeX{{\rm B\kern-.05em{\sc i\kern-.025em b}\kern-.08em
    T\kern-.1667em\lower.7ex\hbox{E}\kern-.125emX}}
\begin{document}

\title{Position Paper: Bounded Alignment: What (Not) To Expect From AGI Agents}

%\author{\IEEEauthorblockN{Anonymous}
%\vspace{10mm}
%}

\author{\IEEEauthorblockN{Ali A. Minai, {\it Senior Member IEEE}}
\IEEEauthorblockA{\textit{Dept. of Electrical and Computer Engineering}\\
\textit{University of Cincinnati}\\
Cincinnati, USA \\
minaiaa@ucmail.uc.edu}
}

\maketitle

\thispagestyle{firstpage}

\begin{abstract}
The issues of AI risk and AI safety are becoming critical as the prospect of artificial general intelligence (AGI) looms larger. The emergence of extremely large and capable generative models has led to alarming predictions and created a stir from boardrooms to legislatures. As a result, AI alignment has emerged as one of the most important areas in AI research. The goal of this position paper is to argue that the currently dominant vision of AGI in the AI and machine learning (AI/ML) community needs to evolve, and that expectations and metrics for its safety must be informed much more by our understanding of the only existing instance of general intelligence, i.e., the intelligence found in animals, and especially in humans. This change in perspective will lead to a more realistic view of the technology, and allow for better policy decisions.
\end{abstract}

\section{Introduction}
\label{introduction}

The most successful AI systems today, such as {\it large language models} (LLMs) \cite{OpenAIGPT4:2023,GeminiTeam:2023,Anthropic_Claude3:2024,Grattafiori_Llama3Herd:2024,DeepSeekV3:2024}, are based on a computationalist, statistical, and decision-theoretic paradigm rather than a biological one. As these systems scale up in size, they are improving their performance in areas such as reasoning \cite{Chen_LLMreasoner:2024,Xiang_LLMreasoning:2025,DeepSeekR1:2025,Liu:2025FoundationAgents}, and becoming more multimodal \cite{Li:2023multimodal,Liang:2023,Yin:2023,Jin:2024,Wang_multimodal:2024}. AI agents \cite{ParkGenerativeAgents:2023,ShavitAIagents:2023,Wiesinger:2025}, including physical ones \cite{Black:2024Pi0,Radosavovic:2024humanoid,Ma:2024VLAsurvey}, are also becoming increasingly capable. With these rapid advances, there is an expectation that powerful systems with {\it artificial general intelligence} (AGI) may soon be at hand. Through all this, there is a general desire that AGI must remain subject to human control and intervention, and must exist only to serve human needs (see, for example, the discussion in \cite{Pace:2019}). There is also great concern that increasingly powerful AGI systems with autonomous agency might pose serious risks, including existential ones \cite{Omohundro:2008,Yudkowsky:2008,Bostrom:2012,Strickland:2023,Carlsmith:2024,Bengio:2025AIrisk}, which has led to a focus on {\it AI alignment}, i.e., making AI systems consistent with human norms and preferences \cite{Leike:2022alignment,Gabriel:2024AIethics}. 

The main position argued in this paper is that: 1) General intelligence should be seen in terms of its archetype: \textit{The intelligence of living agents}; and 2) The goal of building powerful AGI agents is fundamentally inconsistent with the expectation of complete alignment or near-total control of AGI agents by humans {\it even in principle}. The best that can be achieved is {\it bounded alignment}, defined by analogy with \textit{bounded rationality} \cite{Simon:1955} as follows: \textit{The agent's behavior is almost always acceptable -- though not necessarily optimal -- for almost all humans who interact with it or are affected by it}. The degree of alignment expected from AGI should be no more than that expected from well-behaved humans and trained animals.

\section{Why Alignment is Hard}

Broadly, the goal of alignment is to make AI agents safe by constraining them to follow acceptable norms of behavior. However, this goal conceals a great deal of complexity because it is difficult to turn the abstract problem of alignment into a concrete task that can be accomplished computationally. Alignment aims to constrain two things in particular: Goals and behavior (including expression), and to do so by aligning the {\it values} of the agent with human values -- or, at least with human {\it preferences}. A great deal of thinking \cite{Yudkowsky:2008,Christian:2020,Ngo:2025DLalignment} and methodological research \cite{Christiano_PrefRL:2017,Ouyang:2022,Bai:2022RLHF,Bai:2022RLAIF,Rafailov_DPO:2024,Guan:2025DeliberativeAlignment,Qi:2025DeepAlignment} has gone into how this can be achieved. 
However, the task of alignment faces several challenges, including the following.
\newline
\newline
1. \textit{Values are vague}: Values and preferences are extremely hard to define quantitatively and consistently across the entire spectrum of possible behaviors and objectives, and across human differences in culture and ideology \cite{Russell:2019,Gabriel:2024AIethics}. Whose values should the agent align with? And how should it make sure that it acts appropriately in all situations, not just those it was trained in? In fact, ``alignment'' is not quite the right term for the goal of creating safe AGI, since it implies that there is something to align \textit{with}, and no such unique set of values or preferences can be defined rigorously. The values that the agent needs to acquire, instead, are those embodied in a \textit{personal code of ethics} that it applies to all its goals and actions. And, while these values may be universal from the agent's viewpoint, they will still not be universally acceptable to all those it interacts with, or in all contexts. The possibility of a universally aligned, explainable, trustworthy AGI agent is no more realistic than that of a universally aligned, explainable, trustworthy human being.
\newline
\newline
2. \textit{Intelligence is inherently risky}: Attempts to build increasingly powerful AGI agents will necessarily require the agents to have open-ended behavioral complexity, autonomy, self-motivation, life-long learning, and other attributes of general intelligence even if the ultimate goal is only to serve human needs, and these attributes will inevitably give such agents the capacity for inappropriate or dangerous behavior.
\newline
3. \textit{The agent and the environment are both complex dynamical systems}: The most fundamental problem for AGI alignment is the nature of general intelligence itself. Any AGI agent will necessarily be an \textit{autonomous complex adaptive system} (ACAS) operating in an extremely complex dynamic environment. The behavior of the agent will {\it emerge} from its interaction with the environment in the context of its {\it available affordances} and its {\it internal perceptual, cognitive, and motivational states}, which, in turn, will influence the state of the environment in highly complex ways. This creates several problems: 1) It will be impossible to anticipate all the situations that the agent might encounter, and to predict whether the behavior that emerges will be aligned; 2) The values and preferences that underlie this behavior will also be emergent \cite{Omohundro:2008,Mazeika_2025LLMvalues}; 3) Any values and preferences that the agent may initially have could change as it experiences and learns from its complex environment; and 4) Since any AGI agent will have affordances and state spaces very different from those of humans, they will typically not share a common representational framework in which they can agree on values any more than humans do with other animals. 

The essential lesson of these impediments is that the goal of perfect alignment and perfectly safe AGI misunderstands the nature of general intelligence and underestimates its complexity.

\section{S-Agents and P-Agents}

Since its inception, there has been a conceptual paradox at the core of the AI enterprise: Is the goal of building AI: a) to provide tools to achieve human goals, or b) to create autonomous intelligent entities with their own motivations and goals? The implicit assumption behind most work in AI has been that (a) is the right answer. This is the view motivating everything from the so-called Laws of Robotics proposed by Asimov \cite{Asimov:1950} to the current work on AI safety \cite{Russell:2019,Dalrymple:2024GuaranteedAIsafety,Greenblatt:2024AIcontrol,Bhatt:2025AIcontrol} and alignment through post-training reinforcement learning \cite{Christiano_PrefRL:2017,Ouyang:2022,Bai:2022RLHF,Bai:2022RLAIF,Rafailov_DPO:2024,Guan:2025DeliberativeAlignment,Qi:2025DeepAlignment}. But is it logical to assume that AI systems will remain captives of human goals even as they progress towards achieving general intelligence?
 
Some crucial attributes that we would expect in a submissive agent include obedience, reliability, veracity, honesty, transparency, prosociality, etc. Together, these and other similar attributes may be termed \textit{safety attributes}, or \textit{S-attributes}, and an agent with these S-attributes designated an \textit{S-agent}. While S-attributes would make the agent highly aligned, they would not give it the ability to solve complex real-world problems or perform complex real-world tasks. Even assuming that the agent's only purpose is to serve human needs, if it seeks to satisfy goals across the entire spectrum of human needs and desires, it must confront the reality that: a) No objective can ever be specified by a human so completely that it excludes all harmful possibilities, as demonstrated by the famous {\it paperclip task} \cite{Bostrom:2014}; and b) The world is an extremely complex, unpredictable, nonlinear, nonstationary dynamical system with emergent phenomena, an infinite number of unexpected situations, and an untold number of other active agents -- living and artificial -- with their own goals and behaviors. This means that much of the agent's own behavior must be generated on the fly rather than being pre-planned. While a human user may specify the ultimate goal for an agent at some level, it is up to the agent to come up with the necessary {\it sub-goals} and the \textit{ means} to accomplish them in real time. Indeed, the entire purpose of building an AGI servant would be to free humans from this responsibility.
%The AGI agent must come up with the means and make its own choices, and do so with great intelligence and efficiency in real time.

Meeting this challenge will require the agent to have a set of \textit{performance attributes}, or \textit{P-attributes}, including: 1) \textit{Autonomy} to act without continuous guidance; 2) \textit{Self-motivation} to pursue internally-generated goals; 3) \textit{Creativity} to solve hard problems; 4) \textit{Imagination} to consider hypotheticals and counterfactuals; 5) \textit{Introspection} for making deliberative choices; 6) \textit{Versatility} across a wide range of domains; 7) \textit{Perceptual breadth} to make sense of multimodal real-world situations; 8) \textit{Cognitive depth} to create, store, recall, and use concepts and ideas across a hierarchy from concrete to abstract; 9) \textit{Cognitive control} to attend selectively to internal mental states and make wise choices; 10) \textit{Cognitive agility} to revise plans continuously in real time as needed; 11) \textit{Behavioral flexibility} to adapt behavior in real time as needed; 12) \textit{Exceptional capabilities} to do things that humans cannot do; and 13) \textit{Open-ended, continuous, autonomous, life-long learning} to retain and improve capabilities in a dynamic world. Of course, this list is not exhaustive, but it attempts to encapsulate the things that would give an agent the kind of general intelligence needed to behave intelligently in the real world. An agent with these attributes is termed a \textit{P-agent}. 

It is easy to see that any generally intelligent agent we build to serve even quite specific human needs -- say, a robot restaurant waiter or nursing home attendant -- will need \textit{all} these attributes if it is to be useful. Anything less would be worse than useless because the agent would keep making bad choices, not know what to do in unexpected situations, and require constant instruction or approval. The question at the heart of alignment is whether a P-agent can also be an S-agent.

%Many scenarios can be described where P-attributes would necessarily compromise S-attributes.
An autonomous agent is, by definition, beyond total human monitoring and control. If it is creative, imaginative, and has complex behavioral and mental capabilities, it will be able to generate complex ideas, plans, and strategies, and to execute difficult tasks across many domains, which is exactly what we would like it to do. But these very attributes would, in principle, allow the agent to deceive, disobey, and inflict harm. Indeed, these negative abilities are features, not bugs for any intelligent agent in a complex and dangerous world. But the AGI agent will go beyond that: Its vast computational resources would enable it to out-think humans, and to learn continuously across what could be a lifespan far longer than that of a human. Any ``factory settings'' for good behavior could thus be erased over time by dilution or by deliberate choice. The inescapable dilemma is that the very things that make AGI useful also make it unsafe. There is a safety-utility tradeoff between S-attributes and P-attributes, and an ideal AGI agent will have to compromise on S-attributes to find the right balance, precluding the possibility of perfect alignment. This tradeoff is qualitatively different than the risk-reward tradeoff for other high-risk technologies such as nuclear, biological, and chemical weapons for several reasons: a) Unlike AGI, the underlying science in those cases is clear-cut and the risks are well-understood; b) Those technologies have narrow applications, providing feasible regulatory targets, whereas AGI will be applied to virtually everything; c) Those technologies are passive and require human agency for their use whereas AGI will be autonomous and self-motivated with the ability to act on its own; and d) Even a relatively low level of P-attributes will give an agent the potential to bootstrap itself to ever higher capability levels, thus increasing risk in an open-ended way. 

In summary, building an AGI agent with P-attributes is fundamentally incompatible with perfect alignment. Existing generally intelligent agents, i.e., animals (including humans), demonstrate this limitation, and we do not expect fellow humans, pets, or domesticated animals to be perfectly well-behaved even after extensive training.
Recognizing this dilemma, we should accept that the \textit{alignment problem} will never be ``solved'' completely. The goal should instead be to achieve robust, long-lasting bounded alignment, and that will require a deeper understanding of general intelligence itself.
%They will need socialization, not explicit alignment, and this requires a fresh look at what AGI will ultimately be.

\section{Rethinking AGI}
\label{AGI}

\subsection{General Intelligence}
Defining AGI has proved to be quite challenging due to the breadth of issues involved and an emphasis on computational formalisms \cite{Turing:1950,Goertzel:2014,Morris:2024}. Some have defined it as AI that can perform almost all economically useful tasks that humans can perform at a human level of competence or better, which is a rather anthropocentric way of looking at intelligence. This position paper argues for a broader, more agent-centric and biologically grounded view, defining general intelligence as follows:
\newline
\newline
\textbf{General Intelligence:} \textit{The ability of an autonomous agent to exploit its environment productively, actively, creatively, and opportunistically in \textit{almost all} the situations it encounters in its native environment -- including extremely novel ones.}
\newline
\newline
Thus, general intelligence is not defined as the ability of the agent to perform certain tasks to serve its users, but as its ability to exploit the possibilities of its environment pervasively \textit{on its own behalf}. 
A key point is that the utility function implicit in this exploitation may be arbitrarily complex and inaccessible even to the agent itself. Thus, the potential behavior that emerges in the course of the exploitation is open-ended and unknowable \textit{a priori}. It is possible to be generally intelligent in an infinite number of ways for different agents in different environments. 

\subsection{Natural General Intelligence}

To understand how general intelligence arises and how it works, it is useful to look at the only general intelligence that already exists, i.e. the {\it natural general intelligence} (NGI) of animals with brains. The  intelligence of NGI agents emerges from the interaction between their \textit{specific form} and their environment, and as they experience that environment, they reorganize themselves continually in ways that correspond to acquiring new knowledge and the emergence of new skills. A generally intelligent agent embedded in a complex dynamic environment is necessarily one that learns and evolves throughout its life; otherwise it would lose the ability to keep exploiting its environment. Here, the term ``specific form'' is used deliberately instead of the more common term ``embodiment'' because the latter has come to refer mainly to an agent's external, macroscopic form, whereas intelligence arises from the configuration of structures and processes from the macro to the cellular and molecular level in physical agents, and across a similarly wide range of scales in virtual agents.

This definition of general intelligence extends the concept of AGI to all kinds of broadly competent artificial intelligent agents -- present and future -- and allows research on issues within AI to connect with the increasingly unified understanding of intelligence and agency in living systems from the simplest to the most complex \cite{Mitchell:2023,Jacobsen:2024,Seifert:2024}.

\subsection{Intelligent Agents}

Intelligence is ultimately a basis for generating behavior in the agent's specific environment. The agent's sensors and effectors represent its \textit{external embodiment} (or body), but a particular \textit{mental architecture} is also necessary. The entire evolutionary process can be seen as the co-evolution of increasingly complex external embodiments and mental architectures, resulting in increasingly complex general intelligence. Several key components are essential to the functional configuration of an intelligent autonomous agent's mental architecture, and they are briefly discussed below. Of these, the first four -- perception, cognition, behavior, and drives -- form the core, while the other three represent more general aspects of intelligence. As we think of AGI agents, it is useful to ask how they map on to this framework.

\textbf{Perception:} A huge amount of redundant multimodal sensory information flows into the agent in real-time. It must be processed into a form that gives the agent a consistent, integrated, and actionable view of its environment. This is the role of {\it perception}. 
The result of the perceptual process is the creation of a {\it perceptual space} in terms of which the agent experiences its world. 

\textbf{Behavior:} Formal decision frameworks such as reinforcement learning (RL) typically model behavior as sequences of individual actions within an \textit{action space} \cite{Sutton:1998}, but behavior in agents with general intelligence is far more complex in several ways: It is real-time, continuous, high-dimensional \cite{Bernstein:1967,Morasso:2022}, hierarchically organized \cite{Merel:2019}, and multiscale in space and time \cite{Dehghani:2018,Kays:2023}.
Based on these factors, it is more appropriate to think of an intelligent agent as operating in an extremely complex, multi-level, multiscale, open-ended, self-organizing \textit{behavioral space} rather than a predefined discrete or continuous action space.

\textbf{Cognition:} Linking perception to behavior is arguably the primary competence of an agent, and cognition is the process mediating this linkage -- creating the so-called \textit{sense-think-act cycle}. Conceptually, it plays the role of a very complex and deep, multi-recurrent ``hidden network'' linking the perceptual space to the behavioral space bidirectionally. Like perception and behavior, cognition too is an active process with complex, multi-scale dynamics  rather than the set of feed-forward transformations found in most deep neural networks today. Most importantly, it provides a rich, hierarchical representation of the world at levels ranging from concrete features grounded in sensorimotor experience to abstract ones \cite{Kiefer:2012}. The depth of this hierarchy -- the agent's \textit{cognitive space} -- plays a central role in determining its degree of intelligence.

\textbf{Intrinsic Drives:} Unlike passive AI models such as LLMs, animals possess internal drives including hunger, thirst, the mating urge, curiosity, emotions, etc., to motivate their behavior. These drives are a critical component of any generally intelligent agent because they lie at the root of its agency -- its role as an autonomous, self-motivated actor in the world. An agent without curiosity, the urge to explore and exploit its environment, a sense of self-preservation, the motivation to learn, and the need to achieve its own goals, could not be successful in a complex, dynamic, hazardous, ever-changing world. In any generally intelligent agent, drives form a hierarchy that is rooted in a fundamental set of \textit{primal drives} -- notably self-preservation. These give rise to more specific drives such as risk avoidance, dominance-seeking, etc. In humans, this hierarchy is very deep with complex drives such as creativity, ambition, and a desire for social approval or self-expression. The drives of the agent also change with time due to factors such as aging, injury, change of context, and learning.

\textbf{Affordance Space:} {\it Affordances} are the possibilities for perception and action that an environment offers to the agent. The productive exploitation of the environment by the agent lies in effective use of these affordances, and this is precisely what intelligence enables. The key point is that these affordances are \textit{relative}, not absolute: They are {\it relations} induced emergently between particular aspects of the environment and specific perceptual and behavioral capabilities of the agent \cite{Chemero:2003}, or, in Gibson's memorable phrase, they imply ``the complementarity of the animal and the environment'' \cite{Gibson:1977}. 

Thus, for an agent with a specific form, a specific environment induces a specific but infinite \textit{affordance space} that circumscribes the agent's perceptual and behavioral possibilities as well as the time-varying, context-dependent \textit{availability} of these possibilities. While the agent's mental experience of its world is defined by its four core components, it is the affordance space that implicitly generates its self-perception of agency in that world. Affordances may be discovered by the agent through experience, lost due to injury, or become available through practice or via the use of tools.

\textbf{World Model:} The concept of a world model is widely used in AI to refer to an agent's cognitive model for prediction and planning in its environment \cite{Ha:2018,Freeman:2019,Zhang:2021,Friston:2021}, often in a reinforcement learning framework \cite{Sutton:1998,Hafner_Dreamer:2024}. A more general definition comes from LeCun \cite{LeCun:2022}, who equates world models with ``common sense'' that ``can be seen as a collection of \textit{models of the world} that can tell an agent what is likely, what is plausible, and what is impossible'' in its world. This, as LeCun points out, is the reason an intelligent agent can apply its intelligence to \textit{everything} in its world, including completely novel situations. Essentially, the world model must represent the \textit{deep structure} and \textit{causal relationships} in the world it models at a level commensurate with the agent's affordances. Importantly, the world model is not a passive oracle -- a resource to be queried by a higher-level decision-maker as is assumed in RL -- but a deep, richly multi-scale, real-time \textit{active mediator} between perception and behavior, continuously generating hypotheses, predictions, and recommendations in the context of the agent's internal state and affordances as well as the changing state of the environment, which has its own complex dynamics.
%The world model can be seen as providing an extremely large number of adaptive and dynamic internal degrees-of-freedom for the agent to generate behavior \textit{at multiple time-scales}, ranging from milliseconds to years. 
%The quality of an agent's intelligence in its world depends fundamentally on the quality of its world model, which, in turn, depends on the agent's cognitive space. 
The implicit goal of life-long learning for the agent is the improvement of its world model to expand its ability to exploit more of its affordances, and possibly recognize new ones.

\textbf{Memory:} A critical element in any intelligent agent is memory, i.e., the ability to retain information from past experiences to influence future behavior. Memory in biological agents is pervasive through all aspects of mental function. Broadly, it can be divided into types such as short-term \textit{working memory} for storage of currently needed information \cite{Baddeley:2003}; long-term \textit{declarative memory}, including \textit{episodic} and \textit{semantic memory}; and \textit{implicit memory}, which includes \textit{procedural} and \textit{emotional memory} among other types. As the locus of learning and the basis of behavior, memory of all types is especially important for a functioning world model. It is also the key integrator of the perceptual, cognitive, motivational, and behavioral components of the agent, forming representations across modalities and at every hierarchical level. In animals, memory involves both the brain and the body, and this will be the case for embodied AGI agents as well.  

To summarize, any AGI agent will need to have all the components of intelligence listed above -- explicitly or implicitly. Whatever its specific form -- physical or virtual -- it will have sensors and effectors, and will interact with its environment to induce an affordance space. It will learn a world model grounded in its perceptual, behavioral, and affordance spaces in the context of its internal state and drives. Any such agent will have preferences across all these modalities, and these will implicitly define its values, biasing how it perceives the world and how it chooses to act in it.

\section{The Need for A Paradigm Shift}

\subsection{Contrast with the Current AI Paradigm}

The model of AGI proposed above is quite different from the dominant paradigm in machine learning-based AI today which is focused on computational mechanisms, optimizing specific objectives, accomplishing specific tasks, and building towards a more general version of AI by adding tasks and modalities gradually while maintaining the same general paradigms (typically, autoregressive generation). 
The architectures and learning algorithms of these models are regular, stereotypical, and fundamentally different from those seen in biological agents, with \textit{generic architectures} such as stacked attention layers, supervised (or self-supervised) learning using gradient descent, and severely limited, narrow objectives such as next token generation or pixel updating. The sensory inputs of multimodal LLMs \cite{Li:2023multimodal,Liang:2023,Yin:2023,Jin:2024,Wang_multimodal:2024} consist of token sequences representing text, numerical data, code, images, and audio, so their perceptual affordances are limited to these modalities. It can be argued that internal embeddings generated in the early layers of deep neural networks such as LLMs or CNNs correspond to perceptual space representations, and the later layers can be seen as performing cognitive transformations based on a cognitive hierarchy implicit in their weights, producing a cognitive representation in the final embedding layer. This immediately raises the question of whether even very large embedding vector spaces have anywhere near the richness of the perceptual and cognitive spaces of a human or even a fish. While there is a growing body of work showing some correspondence between representations found by deep neural networks and those in the brain \cite{Walker:2019InceptionLoops,Hosseini:2024UniversalRepresentations,Goldstein:2024}, these are typically confined to single modalities.

The behavioral spaces of LLMs are typically limited to token generation. Sequences of such actions correspond to generated text or code, which can then be interpreted by other tools to create images or take other actions.  Their working memory consists of the context buffer and the signals carried by the residual path in the network, and is thus quite superficial with no ability to store discrete internal states. Their long-term memory is distributed across their weights, with no separation between semantic and procedural memory. Classic LLMs do not have episodic memory, though OpenAI recently added a feature allowing GPT-4 to search through previous conversations. LLMs generate their behavior in a ``ballistic'' way with no internal recurrence, introspection, or deliberation. Only very recently have there been rather simplistic attempts to add in deliberative, System 2 mechanisms \cite{Kahneman:2011}, using post-training reinforcement learning and inference-time computation \cite{Chen_LLMreasoner:2024,Xiang_LLMreasoning:2025,Guan:2025DeliberativeAlignment,DeepSeekR1:2025} or latent-space deliberation \cite{Liu_Deliberation:2024}. 

LLMs can be seen as learning a world model, but their ``world'' is the one defined by text, code, numerical data, images, etc., not the physical world. Thus, they internalize the causal and logical relationships underlying grammar, syntax, semantics, and, to some degree, image structure. While such models have some utility for exploration and decision-making in the real world as well \cite{Xie:2024_LLMworldModels}, this is based on second-hand ``book learning'' rather than direct experience, and thus often fails on simple tasks. At best, such models represent an important first step towards building AGI agents.

One of the most important developments in AI recently is a surge of new work on building AI agents based on LLMs \cite{ParkGenerativeAgents:2023,ShavitAIagents:2023,Wiesinger:2025}. These models leverage generative AI capabilities to create a degree of agency, and greatly expand the behavioral space by adding various effectors such as code interpreters and embodied manipulators. This has come to be termed {\it agentic AI}. In practice, most of these agents are still focused on specific tasks or domains, though it is possible to define a hierarchy from specificity to generality \cite{ShavitAIagents:2023}. Recently, there have also been proposals for more open-ended, creative agents that can define their own environments, goals, and tasks \cite{FaldorOMNI-EPIC:2024}. A key innovation is the development of \textit{vision-language-action models} (VLAs) \cite{Black:2024Pi0,Radosavovic:2024humanoid,Ma:2024VLAsurvey}, which integrate LLMs and action prediction models to allow fairly complex behavior learning in embodied agents.

As AI moves towards AGI, agents will \textit{inevitably} have to become more natural and incorporate more and more features seen in biological agents \cite{Liu:2025FoundationAgents}, but may do so in radically non-biological ways.  These agents will differ greatly from those we see today. They will be active, autonomous, self-motivated, introspective, and deliberative. They will have affective states and extremely complex world models. Their behavior spaces too will be extremely rich and complex, but, depending on their particular forms, could be defined on affordance spaces very different from those of humans, just as the behavioral spaces of other animals are. These agents will thus be \textit{alternative intelligences} -- even {\it alien intelligences} \cite{Harari:2024} -- that may inhabit worlds and have thoughts and values incomprehensible to humans, thus posing arbitrary and open-ended risks \cite{Bengio:2025AIrisk}. AGI alignment must focus primarily on mitigating this risk.

\section{Aligning Alternative Intelligences}

\subsection{Alternative Intelligence}

In 1974, cognitive science pioneer Thomas Nagel published a seminal paper \cite{Nagel:1974} titled, ``What is it like to be a bat?'' The paper argued that, given a bat's specific embodiment and set of mental and physical capacities (i.e., affordances), it is impossible for a human to imagine how the bat experiences the world inwardly \textit{as a bat} (rather than as a human simulation of a bat). 
Indeed, every agent with a specific form experiences the world in terms of it own perceptual space, world model, motivational states, and affordances. This poses a fundamental challenge for agents in understanding the experiences of other agents, but the difficulty depends strongly on the types of agents involved.

It is relatively easy for us humans to understand the experiences of fellow humans because of shared biology, shared form and affordances (with minor variations), shared experiences, and the brain's mirror system \cite{Rizzolati:2004MirrorSystem}, which allows an individual to represent an observed action by another person in a  representational space shared with their own actions. Thus, humans have high-quality {\it theories of mind} for other humans. This is much more difficult when trying to understand the mental experiences of other animals such as Nagel's bat, but even so, these animals -- from cats and dogs to bats, bees, and octopuses -- share many things with us, including the same fundamental biology; many homologous genes, tissues, and organs; many of the same needs and hazards; and, crucially, similar drives embedded in all animals by evolution \cite{Mitchell:2023}. Thus, we can at least describe and comprehend their behaviors and guess their motivations, if only in human terms. When a bird flies away as we approach, or a bee stings us if we swat at it, we may not share their mental experiences, but we do understand their motivations because we too run away from danger and fight back when attacked. Similarly, we comprehend the motivations of a spider spinning a web, a mole digging a burrow, and a tiger stalking prey. They are fellow animals -- even distant relatives -- and are doing things that we can identify with as humans. We also have the technical language to describe these behaviors at many levels using the frameworks developed by biologists, neuroscientists, and psychologists, though decades of research has still only given us limited insight into the minds of intensively-studied animals such as rats and chimpanzees. AGI agents will be far more \textit{inscrutable}. 

Unlike every other intelligent agent we know, AGI agents will not be biological. They may be virtual or embodied, but, in their material or electronic substrates, their forms, their information processing mechanisms, their sensors and effectors, their affordances, and in terms of their metabolic needs and vulnerabilities, they could be completely different. They will not face the same hazards (e.g., particular chemical toxicities and injuries), nor experience animal-like kinship ties or go through the experience of aging and death on the same timescale as most animals. They will potentially have memory capacities and information processing capabilities far exceeding those of humans, direct access to cyberspace, and, most importantly, the ability to easily replicate their minds, if not their bodies, and to modify and improve them rapidly. With all these differences, AGI agents will inhabit worlds completely different from those of any animal. Indeed, they may not even share worlds with each other because of their radically diverse forms \cite{Yudkowsky:2008}. How, then, could such diverse alternative intelligences possibly be aligned with humans? And are the current methods being used for alignment suitable for this?

\subsection{Current Approaches}

As discussed above, in the context of AGI safety, \textit{values}, refers to a systematic set of implicit principles that modulate the motivations, goals, and behavioral preferences of an agent globally. 
These, in turn, depend on how the agent sees the world (perception), its internal motivations (drives), its causal understanding of the situation (world model), and its available behavioral possibilities (behavioral affordances). Thus, values in an aligned AGI agent must be implicit in \textit{all }components of the agent's mental processing and shape the flow of information through all of them, biasing the agent towards having an \textit{aligned disposition} -- if only in a bounded sense. The issue is whether current approaches to alignment can achieve this, and whether there is a better way.

The dominant practice for value alignment in current AI agents is to take pretrained base models and teach them to prefer acceptable responses through reinforcement learning and/or fine-tuning
\cite{Christiano_PrefRL:2017,Ouyang:2022,Bai:2022RLHF,Bai:2022RLAIF,Rafailov_DPO:2024,Guan:2025DeliberativeAlignment}, thus effectively biasing their weights to implicitly embed human preferences. While sound in principle and consistent with the idea of values as inherent biases, these approaches have serious practical limitations. They are very expensive to scale, rely on uncertain inductive reward models, and assume that the agent is under human control and subject to human training or observation, which will not always be the case with AGI. Another problem is that the training process can lead to the learning of emergent values that may be inappropriate but hard to detect using benchmarks. Mazeika et al. recently reported that, with scaling, LLMs aligned using current methods converge emergently to coherent value systems that can be ``problematic and often ``shocking'', including ``cases where AIs value themselves over humans and are anti-aligned with specific individuals'' \cite{Mazeika_2025LLMvalues} -- presumably because of latent biases in the data. They propose a principled {\it utility engineering} framework for probing emergent values in LLMs, and using it to train them towards a value system obtained from a simulated citizen community. While insightful, such an approach in its current form is applicable mainly to LLM-like systems and cannot work for AGI agents that are autonomous and are learning continuously on their own outside of human control. Also, the method depends fundamentally on the probing protocol used to estimate the LLM's values, which is likely to miss latent values that the utility engineers failed to imagine. The same applies to the Constitutional AI principle used by Anthropic \cite{Bai:2022RLAIF}, where the constitution in which values are grounded is necessarily limited by the imagination of its authors. 

The main problem with most current alignment methods, however, is their logical structure, where a ``wild'' but highly capable base model is trained on huge amounts of data, and then ``civilized'' using feedback from trained evaluation models or self-generated data. This is like trying to civilize an adult human who was raised by wolves -- actually worse, because the wild model, unlike a wolf-man, already has vast knowledge and generative capabilities. It is possible -- even likely -- that this post-facto alignment will leave a lot of harmful atavistic tendencies latent in the agent, which may emerge unexpectedly in the field. Thus, the alignment produced by current methods is typically brittle because it is superficial -- a thin veneer of civility over a wild, toxic nature that can be unmasked by jail-breaking methods \cite{Zou:2023,Anil:2024_JailBreak}. 

An alternative approach to making AI agents safer is to not assume complete alignment, but to detect misalignment and to mitigate it. A useful tool for this is {\it mechanistic analysis}, which tries to look at the emergent internal representations in deep models, and to determine if they are meaningful, interpretable, and robust \cite{Tenney:2019,Hosseini:2023,Sucholutsky:2024,Huben:2024SAE}. This can potentially be used to explain how the model is making its decisions, and thus validate its acceptability. While very useful for understanding how deep networks work, this approach can only provide indirect and incomplete information on alignment. It is also likely to become infeasible in extremely large models with complex internal dynamics, emergent behaviors, self-motivation, and autonomous continual learning.

Another mitigation-based approach is to use strategies involving continuous monitoring, assessment, and control \cite{Dalrymple:2024GuaranteedAIsafety,Greenblatt:2024AIcontrol,Bhatt:2025AIcontrol}. However, these methods still rely on human anticipation of all possible failure modes and detection of failures using explicit observers and built-in verifiers. Thus, they too are unlikely to scale to AGI that is capable of out-thinking humans, or to be useful for agents operating in the real world beyond constant human supervision and intervention.

Given all these issues, how can robust bounded alignment be achieved in AGI agents? The next section proposes a possible approach for this.

\section{A Natural Approach to Alignment}

\subsection{Principles for Robust Bounded Alignment}

The argument made in this paper is that the following principles can serve as the basis of building AGI agents with the kind of robust bounded alignment seen in humans and trained animals: 

\begin{enumerate}
\item AGI agents should be designed in ways that allow them and humans to have adequate theories of mind for each other. These can then serve as the basis of mutual communication, comprehension, cooperation, and instruction \cite{Williams:2022AI_ToM,Gupta:2025COHUMAIN}, leading naturally to easier alignment and greater likelihood of \textit{corrigibility} (tolerance for instruction and correction) \cite{Soares:2015Corrigibility}. 
\item AGI agents should be given innate characteristics that make them inherently amenable to alignment, possibly including immutable, built-in features that do not compromise P-attributes too much but limit or expose misalignment before it becomes dangerous.
\item Ethical principles should be part of the learning regimen of AGI agents from the beginning rather than being imposed after a base model is trained, thus embedding values deeply into the fabric of the agent's mind.
\end{enumerate}

While there may be many ways to instantiate these principles, an obvious approach is to make AGI agents more like biological agents in their functional architecture, information processing mechanisms, representational frameworks, and drives. This will make them intrinsically less alien, and thus allow them to be analyzed,  understood, and trained in more familiar ways, leading to greater safety and trust. It would also allow insights from biology, neuroscience, cognitive science, developmental psychology, and even economics and philosophy, to be applied more readily to AGI agents. For example, this could make it easier to define innate alignability traits by analogy with humans. 

The idea of building a more brain-like AI is a central tenet of the emerging field of \textit{NeuroAI} \cite{Zador:2023NeuroAI}. A very detailed and insightful study of how NeuroAI methods can be used explicitly to enhance AI safety has recently been published by Mineault et al. \cite{Mineault:2025_NeuroAISafety}, and the approach deserves serious consideration. While the methods proposed in the study are theoretically compelling, they seem rather ambitious for the current state-of-the art in both neuroscience and machine learning, e.g., building detailed embodied digital twins of animals and humans, building foundation models for the brain, and fine-tuning AI systems on large amounts of data from the brain. These are becoming feasible for simpler animals like the fruit fly \cite{Iwasaki:2025DrosophilaTwin}, but scaling them up to humans -- or even rodents -- will be prohibitively difficult (though see \cite{Mathis:2024BrainDecoding,Zhang:2025NeuralEncoderDecoder}). However, the \textit{vision} behind NeuroAI can be applied at a more abstract level to constrain AGI agents to be more \textit{congruent} with natural ones, allowing the agents and humans to have theories of mind for each other and lead naturally to greater alignability. A key point here is that, in making AGI agents more natural, it will be important to look not only at brains and bodies, but also at the evolutionary and developmental processes that make natural intelligent agents what they are in their nature -- an approach that may be called {\it EvoDevoNeuroAI} or {\it BioAI} rather than just NeuroAI \cite{Minai:2024DeepIntelligence}.  

\subsection{Developmental Value Learning}
A key component of this approach from the alignment perspective is to use a more developmental method for value learning. If the goal is to make AGI agents robustly aligned, this alignment must become part of their character, not something added as an afterthought on a wild underlying disposition or provided in a reference document to tell the agent right from wrong. Agents must be given a \textit{conscience}, and to do so, we should look at how humans acquire theirs. The answer is that humans learn values in stages from infancy to adulthood. At each stage, they learn values at a level commensurate with their mental and behavioral capabilities, and these values continue to constrain their choices even as their capabilities grow, e.g., an individual who learns to avoid tantrums as a child is likely to become a more responsible and thoughtful adult. Even a dog or a cat must be house-broken early before their behavioral choices become too complex. For example, prosocial behavior -- a key component of alignment -- emerges developmentally in humans from infancy to adulthood \cite{Jensen:2014_Prosociality}, becoming more complex as the individual's capacities grow. The same is true of other character traits. This principle should also be used for AGI agents, training agents in stages of increasing perceptual, cognitive, and behavioral complexity, and integrating value learning at an appropriate level into the training process at all stages. The goal is to embed the trained values so deeply and inextricably into the agent's mind that violating or erasing them becomes mentally intolerable for the agent. At the same time, it should be recognized that even the deep alignment produced by this process may fail under stress, or as a result of toxic influences, as it can in humans.

\section{Conclusion}

It is likely, if not certain, that artificial agents with significant general intelligence will become feasible in the near future. The risks posed by such agents have been discussed extensively, but much of the recent methodological and policy discussion focuses on large, centralized systems hosted in datacenters controlled by humans -- hence the emphasis on continuous testing, detection, control, and updating. Apart from the obvious ``Skynet risk'' such systems pose if left unconstrained, it just seems likelier that future AGI agents will be smaller, autonomous, self-contained entities loose in the physical world and cyberspace -- millions (or billions) of extremely diverse new non-biological living entities added to the world's ecosystem. They may connect to large AI models in the cloud like humans access the Internet to enhance their cognitive power, but they will have their independent minds. Expecting human testing and control to keep them in line will not be feasible, and adding kill-switch mechanisms will be both counterproductive and futile. Unlike all complex artifacts mass produced through engineering, e.g., computers and aircraft, factory settings and tests will be no guarantee of future behavior because the agents will adapt continuously, interact with other human and AGI agents, and change over time. These will be the first major human technology to which the classic engineering paradigm will be fundamentally inapplicable.

Given all this, some of the questions the AI community should think about to make the AGI of the future as safe as possible are the following: 1) How should alignment be defined and analyzed in agents with alternative intelligences featuring radically different affordance spaces and internal representations than those of humans? 2) What system architectures, computational mechanisms, and learning processes should be used in AGI agents to make them more inherently alignable? 3) Can we define a finite (if growing) set of well-understood canonical models for AGI agents -- like species or genera in animals -- rather than an uncontrolled diversity of distinct models? 4) How can we enable AGI agents and humans to have viable theories of mind for each other, and is an LLM-like approach appropriate for this? 5) What characteristics must explicitly be avoided in building AGI models, and why? and 6) What useful innate values can feasibly be built into AGI agents, and how?

In the end, AGI agents will not be completely alignable with humans -- and perhaps not even with each other. The best that can be hoped for is to have mechanisms for being well-behaved by mutual consent, which is the assumption we make naturally in dealing with other humans and animals. Even so, any alignment that is achieved could still fail as the agent or its environment changes over time, so living with AGI will be a continuous process of learning and accommodation between humans and AGI agents. And if AGI agents do become significantly more intelligent than humans, it may no more be possible for us to constrain their attitudes than chimpanzees can constrain ours. They may turn out to be benevolent towards humans, or not. We must either accept to live with this risk or hope that true AGI never comes to pass.

\balance

%\bibliographystyle{unsrt}
%\bibliography{newbib}

\end{document}